%% file: egpaper_final.tex
\documentclass[10pt,twocolumn,letterpaper]{article}

\usepackage{cvpr}
\usepackage{times}
\usepackage{epsfig}
\usepackage{graphicx}
\usepackage{amsmath}
\usepackage{amssymb}
\usepackage{times}
\usepackage{latexsym}
\usepackage{booktabs}

\cvprfinalcopy 

\def\cvprPaperID{****} 

\ifcvprfinal\fi
\begin{document}

\title{Estimation of Vehicular Velocity based on Non-Intrusive stereo camera}

\author{Bikram Adhikari\\
{\tt\small badhika5@gmu.edu}
\and
Prabin Bhandari\\
{\tt\small pbhanda2@gmu.edu}
}

\maketitle
\thispagestyle{empty}

\begin{abstract}
   The paper presents a modular approach for estimation of leading vehicle's velocity based on non-intrusive stereo camera where SiamMask is used for leading vehicle tracking, Kernel Density estimate (KDE) is used to smooth the distance prediction from a disparity map and LightGBM is used for leading vehicle velocity estimation.
   Our approach yields RMSE of 0.416 which outperforms the baseline RMSE of 0.582 for the SUBARU Image Recognition Challenge\cite{signate}. 
\end{abstract}

\input{introduction.tex}
\input{problem_formulation.tex}
\input{approach.tex}
\input{results.tex}
\input{relatedwork.tex}
\input{conclusion.tex}

{\small
\bibliographystyle{ieee}
\bibliography{egpaper_final}
}

\end{document}

%% file: introduction.tex
\section{Introduction} \label{sec:introduction}

In the past decade, we have seen a lot of change in the automobile industry and the technological shift from stick-shift manual driving to semi-automatic and automatic driving. This technological advancement also brought better safety for the driver, better traffic flow, and a unique concept of multi-vehicle perception and vehicle-to-vehicle(V2V) communication. V2V communication is shared perception is the main focus for major automobile chains moving forward and one of the important features to be shared and inferred is vehicular speed. Vehicular speed and distance perception of the leading and/or the surrounding vehicle can be useful in efficient navigation and also in the estimation of time-sensitive features like time to collide, thus further improving the automatic navigation and road safety in Intelligent Transport Systems. \\
Based on the data that we have, we can estimate the vehicular speed by either the Intrusive sensors-based method or the Non-Intrusive sensors-based method. Intrusive methods are precise in the speed estimation but rely heaving on expensive instruments like Inductive Loop Detectors, these require a heavy investment and engineering to adapt the machines in the vehicle in a non-intrusive manner. The non-intrusive approaches are however cheaper and easier to manage. These non-intrusive methods are further divided based on the frequency and nature of data into Active and passive sensor-based methods. Active methods rely on sensors like RADAR and LIDAR to continuously transmit signals reflected from the environment and help in estimating the vehicular speed based on the time to reflect and distance measure. The problem with these sensors is that they are difficult to manage and infer, costly (not as much as intrusive), and highly subjective to noise. Passive non-intrusive sensors are like cameras that collect the data which can be processed further to estimate the speed. These are comparatively easier to manage and cost-efficient at the same time and are precise enough to assist with the time-sensitive decision-making required in Intelligent Transport System. \\
\begin{figure}[t]
    \centering
    \includegraphics[width=0.48\textwidth]{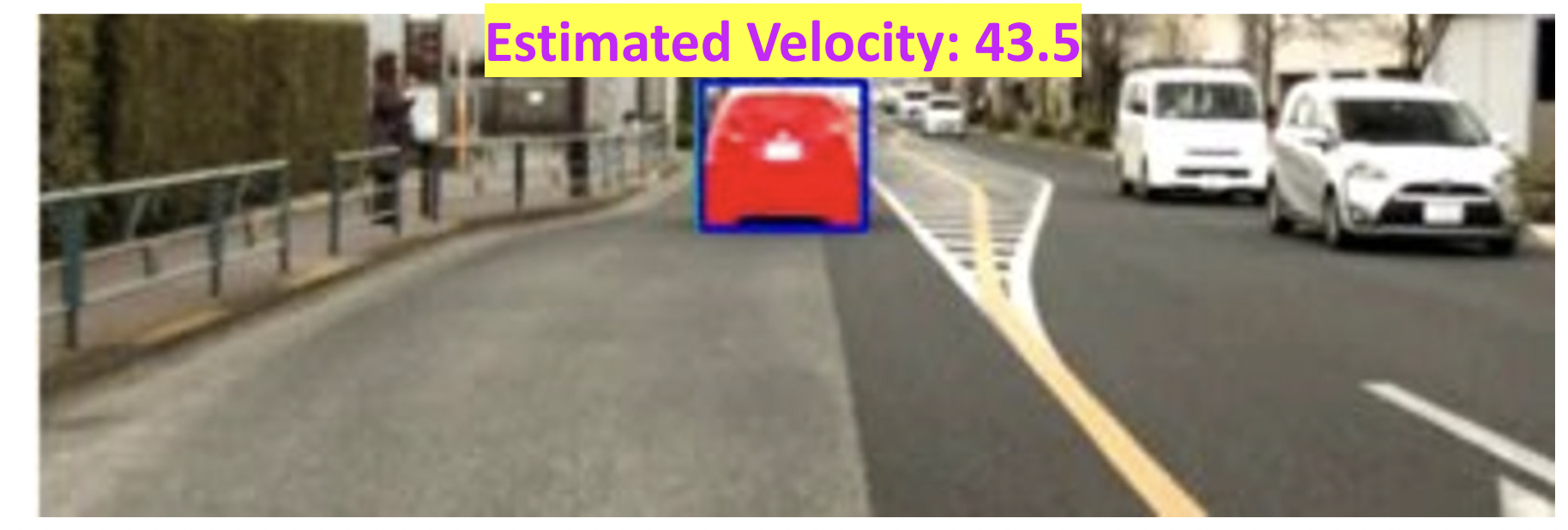}
    \caption{Estimation of leading vehicle's velocity. }
    \label{fig:img_res}
\end{figure}

This research focuses on non-intrusive stereo-camera-based vehicular speed estimation. The problem presented by SUBARU,  a leading automobile company from Japan, we analyze the stereo-video footage from the company's Road-safety and preventive technology named EyeSight. Initially planned as a hardware-specific approach, the company later decided to release the video footage to the general public to use innovative computer vision and machine learning approaches for vehicular speed estimation. The link to the competition webpage is
{\color{red}https://signate.jp/competitions/657}\\  
After this introduction, the paper follows the following order. We start with Problem Description where we discuss the SUBARU competition in detail, the dataset and the features available and our approach of sub-dividing the problem into three sub-problems of progressive leading vehicle tracking, distance estimation from disparity map and leading vehicle velocity estimation. The third section discuss our approach of tackling these three problems. In the forth section we present our Results with SiamMask for progressive vehicle tracking and LightGBM for velocity estimation. Our model was able to perform better than the benchmark by \textbf{0.173 RMSE score }. We finally discuss the Related work in the vehicular speed estimation and our motivation for the approach in this project in Section 5 and we conclude with our final observations and future conclusion. 

%% file: problem_formulation.tex
\section{Problem Description} \label{sec:problem_description}

Based on the dataset provided by the SUBARU Image Recognition Challenge\cite{signate}, we break down the problem of vehicular velocity estimation into three different tasks.
The first task relates to progressive leading vehicle tracking which is followed by distance estimation from the disparity map and the final task is velocity estimation.
We explain each of the tasks along with the dataset below.

\subsection{Dataset}

The dataset consists of video data and annotations broken down into train and test sets.
Video data consists of the right camera video images and left camera video images shot at 10 frames per second (fps) along with their disparity map.
The right camera video images also consist of the rectangular coordinates of the vehicle ahead.
Annotations are provided in JSON format which includes the velocity and steering angle of the vehicle in which the stereo cameras are mounted as well as the velocity of the leading vehicle and the distance between the vehicles.
There are a total of 737 scenes in the train data and 239 scenes in the test data where there are about 100-200 frames per scene.
Test data differs from train data in regards to that the annotations for the test data do not include the velocity of the leading vehicle, the distance between the vehicles and the rectangular coordinates of the vehicle ahead are given only in the first frame.

\subsection{Progressive leading vehicle tracking}

We identify progressive leading vehicle tracking as the first task that needs to be done.
Since we have the rectangular coordinates of the vehicle ahead for the first frame only in the test data, we must have a way to track or regress these rectangular coordinates for subsequent frames.
This problem can be formulated as an object detection problem for each frame but we believe that object detectors have large compute resource requirements and for time-sensitive problems, a better approach would be to do object detection in the first frame through an object detector and then regress the object detection to subsequent frames.
We believe that this is the reason for the competition to provide the rectangular coordinates of the vehicle ahead for the first frame only.

\subsection{Distance estimation from disparity map}

Disparity maps refer to pixel differences or motion between a pair of stereo images.
By matching each pixel in the left image with its corresponding pixel in the right image of a pair of stereo images, we calculate the distance between each matching pair of pixels and represent these distance values as an intensity image to generate a disparity map.
Since the dataset already has the disparity maps, the next task is to map the rectangular coordinates of the vehicle ahead to the disparity map and calculate the distance between the vehicles using the disparity map.

\subsection{Leading vehicle velocity estimation}

Since we have a distance between the vehicle in which the stereo cameras are mounted and the leading vehicle, the final task will be to calculate the velocity of the leading vehicle. With the distance estimation from the previous subtask, we can use the difference in distance given the time (from the video fps), to calculate the relative velocity of the moving vehicle. In theory, by adding the relative velocity to the given velocity of our own vehicle, we should be able to calculate the velocity of the leading vehicle at a given time. Further investigating the problem, we can only calculate the velocity given the difference in time after the second frame as we don't know the position of the vehicle before the first frame. So, another task for us is to estimate the velocity or distance of the leading vehicle before the first frame if we are to proceed with the relative-velocity calculation. 

%% file: approach.tex
\section{Approach} \label{sec:approach}


Based on the task division that we provided in \S \ref{sec:problem_description} for the problem we provide the approach that we take for each task below along with the evaluation metric that we use.

\subsection{Progressive leading vehicle tracking using SiamMask}
\begin{figure*}[h]
    \centering
    \includegraphics[width=\textwidth]{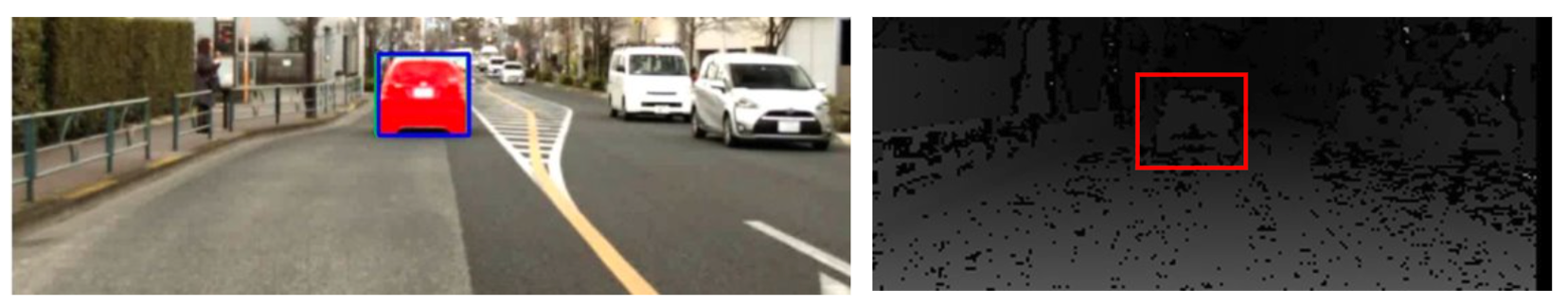}
    \caption{Left image: Shows the bounding box generated for the images from the right stereo camera using the SiamMask model. Right image: Shows the bounding box that is mapped to the disparity map generated from the image of left and right stereo camera. }
    \label{fig:right_disparity}
\end{figure*}

We use SiamMask \cite{wang2018fast} to get the progressive bounding box of the leading vehicle.
SiamMask is an improvement to the fully-convolutional siamese network \cite{bertinetto2016fully} which can be used for real-time object tracking and semi-supervised video object segmentation.
SiamMask relies on a single bounding box initialization and produces online bounding boxes at 55fps.
Given that we have the initial bounding box for the leading vehicle for the first frame, we use this initial bounding box to initialize the SiamMask to produce bounding boxes in subsequent frames.
We train the SiamMask model using  DAVIS 2016 dataset\cite{Perazzi2016} and use the model for downstream object tracking tasks.
The bounding boxes are then mapped to the disparity map.
The process is shwon in Figure \ref{fig:right_disparity}.

\subsection{Distance estimation from disparity map}
\begin{figure}[h]
    \centering
    \includegraphics[width=0.48\textwidth]{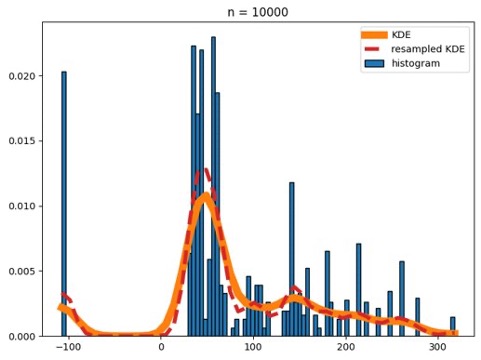}
    \caption{Histogram plot of the per pixel distance generated for the disparity map in figure \ref{fig:right_disparity}, shows that using KDE gives the best estimate of the accurate distance.}
    \label{fig:kde_best}
\end{figure}

Using the bounding box in the disparity map which bounds the leading vehicle, we can calculate the distance from the stereo camera to the leading vehicle using the following formula for each pixel.

\[ distance = {{offset * focal\_length} \over {disparity}}\]

where,
\(offset\) is the separation between the two stereo cameras mounted on the left and right of the vehicle, \(focal\_length\) is the focal length of the stereo cameras mounted on the vehicle, and \(disparity\) refers to the intensity of the pixel in the disparity map.
The overall distance between the vehicle and the leading vehicle is generally calculated as the mode of all the distance measured for each pixel.
Our generated bounding boxes have some inherent problems associated with them for directly using the mode of the distances as the distance between the vehicles.
The bounding boxes do not capture the leading vehicle only, the boxes do capture some of the environment, especially towards the top of the bounding box which can be seen in the left image of figure \ref{fig:right_disparity}.
Another problem would be that based on the distance, the bounding box can be of different sizes.
Closer vehicles will have larger bounding boxes and vehicles further away will have smaller bounding boxes which makes the mode based distance calculation unreliable.
So, We use kernel density estimate (KDE) and resampled KDE to predict the distance.
We got better result using KDE which is shown in \S \ref{sec:results} and figure \ref{fig:kde_best} also shows the superiority of KDE over resampled KDE and mode based approach.

\subsection{Leading vehicle velocity estimation}

Given the distance estimation, we can calculate the velocity based on the relative distance between the vehicles in two subsequent frames but this approach will not work for the first frame.
Further, velocity of a vehicle can be thought of as temporal data as the velocity for a frame is dependent on its velocity a frame ago.
So, we propose to use some regression models to predict the velocity rather than simply doing relative velocity based calculation.
We use calculated velocity from past 18 frames if available as features for our regression model along with the steering angle and velocity of the vehicle in which the stereo cameras are mounted. 
We experimented with Linear Regression , CatBoost, XGboost and lightGBM model for velocity estimation.

\subsection{Evaluation}

Similar to the evaluation metric provided by SUBARU Image Recognition Challenge\cite{signate} , we use Root Mean Squared Error (RMSE) as our evaluation metric. RMSE is calculated for each frame for all the scenes.

%% file: results.tex
\begin{table*}[t]
    \centering
    \begin{tabular}{lllr}
        \toprule
           Vehicle tracking & Distance estimation & Velocity regression   & RMSE \\
        \midrule
        \multicolumn{3}{c}{Baseline score} & 0.582 \\
        \midrule
        SiamMask & Mode & LigthGBM & 0.744 \\
        SiamMask & resampled KDE & LigthGBM & 0.626 \\
        SiamMask & KDE & LigthGBM & \textbf{0.416} \\
        SiamMask & KDE & Relative velocity  & 0.830 \\
        SiamMask & KDE & XGboost & 0.593 \\
        SiamMask & KDE & Logistic Regression & 0.770 \\
        SiamMask & KDE & CatBoost & 0.444 \\

        \bottomrule
    \end{tabular}
    \caption{RMSE obtained for different combinations of solutions for each task. Baseline Score refers to the score provided by SUBARU Image Recognition Challenge\cite{signate}. The lower the value the better the model is. SiamMask for vehicle tracking, KDE for distance estimation and LightGBM for velocity regression produced the best result which is bolded.}
    \label{tab:results}
\end{table*}

\section{Results} \label{sec:results}

The results obtained based on our approaches are shown in table \ref{tab:results}.
We use SiamMask for object tracking in each of our experiments. 
We can see from the second, third and fourth row of the table that using an estimator rather than mode for distance estimation yields better result and KDE produces better result compared to resampled KDE.
We can see from the fifth row of the table that using the relative velocity for between subsequent frames is not a good estimator of the velocity of the leading vehicle.
Bossting algorithms like LightGBM, CatBoost and XGboost performed better than logistic regression for velocity estimation and out of the LightGBM worked the best.
We obtaine the best RMSE score of 0.416 using SiamMask for vehicle tracking, KDE for distance estimation and LightGBM for velocity regression which outperforms the baseline score provided of 0.582.

%% file: relatedwork.tex
\section{Related Work} \label{sec:relatedwork}
With the evolution of computer vision, researchers have put a lot of effort into Single-Object Tracking (SOT) and region proposal generation. 
SOT's objective is to regress the calibration target in the next sequence based on the information from the initial sequence.\cite{liu2021overview}. The SOT trends can be classified into correlation-filter, and Deep-learning based Tracking. \cite{li2018deep}\\
Using a correlation filter to generate a 2-Dimensional similarity map between the candidate and the predicted image, a correlation-filter regresses the position of the target image in the frame based on the maximum value in the similarity map.\cite{li2014scale}\\
However, Correlation-filter based approach is highly sensitive to the candidate initialization, the object's position in the frame, and deformities or object occlusions in the candidate image and the performance degrades if the candidate image is not a good representation of the object of interest. \cite{jiang2019deep}.\\
To overcome the problems related to correlation-filter-based approaches and with deep learning algorithms taking the computer vision field by storm, many deep learning-based trackers \cite{nam2016learning} have been proposed and have also been successful with State-of-the-Art performance. In spite of the performance and results, these models require a large dataset for training and also have a heavy computational overhead for feature extraction and model fine-tuning. So, a novel approach has been introduced in the Visual Tracking paradigm with Deep Siamese networks\cite{koch2015siamese}, which doesn't require explicit feature engineering and a large-computational overhead and training samples for effective similarity mapping.\\
SiamFC\cite{bertinetto2016fully} and SiamRPN\cite{8579033} had SOTA results with the single-object tracking instances but they were limited to two losses associated with the similarity map and the bounding box, so a novel approach of adding a bitwise segmentation mask loss to efficiently estimate the object position was introduced as SiamMask\cite{wang2019fast}. Using the initial bounding frames for the object of interest, SiamMask is able to set new benchmark for Single-object tracking and segmentation.\\
With the change in distance and the change in time, traditionally, physics-based models are used for vehicular speed estimation.\cite{lefevre2014survey} Further smoothing agents Kalman Filters are used to smoothen out the predictions.\cite{kalman1960new}.  However, these models are only perfect for the time-insensitive estimation of vehicular state and are bound by restrictions such as constant velocity or acceleration of the vehicle. They also suffer from prediction rattling and granular results causing further complications in time-sensitive predictions.\\
So, Deep learning based velocity estimation have been used widley for accurate and precise prediction of the vehicular speed.\cite{kim2017probabilistic}\\
So, Deep learning based velocity estimation have been used widley for accurate and precise prediction of the vehicular speed. One such deep-learning based regression model is LigthGBM \cite{ke2017lightgbm} which used Gradient based One-Side Sampling (GOSS) and Bin-bundling of features to converge faster while retaining the efficiency. 

%% file: conclusion.tex
\section{Conclusion} \label{sec:conclusion}
Our results show that using state-of-the-art models for various subtasks rather than in end-to-end settings results in better estimates of the leading vehicle's velocity which helped us to out perform the baseline score.
For improvement to our approach, future research can focus on:
\begin{itemize}
    \item Finetuning the SiamMask model on the specific dataset provide by the competition.
    \item Exploring better techniques for handling false leading vehicle tracking.
\end{itemize}